\documentclass[letterpaper, 10 pt, conference]{ieeeconf}  

\IEEEoverridecommandlockouts   

\usepackage{graphics} 
\usepackage{epsfig} 
\usepackage{amsmath} 
\usepackage{amssymb}  
\usepackage{gensymb}
\usepackage{multirow}
\usepackage{siunitx}
\usepackage{caption}
\usepackage{titlesec}

\usepackage{color} 

\title{\LARGE \bf
mCLARI: a shape-morphing insect-scale robot capable of omnidirectional terrain-adaptive locomotion in laterally confined spaces
}

\author{Heiko Kabutz$^{1}$, Alexander Hedrick$^{1}$, Parker McDonnell$^{1}$, Kaushik Jayaram$^{1,*}$
\thanks{Any opinions, findings, and conclusions or recommendations expressed in this material are those of the authors(s) and do not necessarily reflect the views of the any funding agency. This work is partially funded through grants from the Paul M. Rady Mechanical Engineering Department, US Army research office (ARO) Grant \# W911NF-23-1-0039 and the Meta Foundation (K.J.).}
\thanks{$^{1}$Animal Inspired Movement and Robotics Laboratory, Paul M. Rady Department of Mechanical Engineering, University of Colorado Boulder} 
\thanks{$^{*}${For correspondence, \tt\footnotesize kaushik.jayaram@colorado.edu}}%
}
\begin{document}

\maketitle
\thispagestyle{empty}
\pagestyle{empty}

\begin{abstract}
Soft compliant microrobots have the potential to deliver significant societal impact when deployed in applications such as search and rescue.
In this research we present mCLARI, a body compliant quadrupedal microrobot of $20 mm$ neutral body length and $0.97 g$, improving on its larger predecessor, CLARI. 
This robot has four independently actuated leg modules with 2 degrees of freedom, each driven by piezoelectric actuators. 
The legs are interconnected in a closed kinematic chain via passive body joints, enabling passive body compliance for shape adaptation to external constraints.
Despite scaling its larger predecessor down to \SI{60}{\%} in length and $38\%$ in mass, mCLARI maintains $80\%$ of the actuation power to achieve high agility.
Additionally, we demonstrate the new capability of passively shape-morphing mCLARI -- omnidirectional laterally confined locomotion -- and experimentally quantify its running performance achieving a new unconstrained top speed of \SI{\sim 3}{bodylengths/s} (\SI{60}{\milli ms^{-1}}). 
Leveraging passive body compliance, mCLARI can navigate through narrow spaces with a body compression ratio of up to $1.5\times$ the neutral body shape.

\end{abstract}

\vspace{-3mm}
\section{INTRODUCTION}
One area of significant societal impact that robots have the potential to yet fully realize is in search and rescue operations \cite{yang_grand_2018, drew_multi-agent_2021}. 
To be most effective, robots not only need to traverse over challenging rubble from collapsed structures, but also have the ability to effectively ingress and egress through small crevices in between them \cite{murphy_disaster_2014}. 
To date, snake-like serpentine robots have proven to be most practically successful in these scenarios \cite{xiao_review_2018, liu_review_2021}.
However, the majority of today's legged robots that are beginning to demonstrate robust locomotion capabilities on uncertain terrains are the size of small mammals or larger \cite{hutter_anymal_2017,bledt_mit_2018, guizzo_by_2019}, and therefore still too big for these applications \cite{delmerico_current_2019}.
On the other hand, miniature robots in the past decade \cite{de_rivaz_inverted_2018, doshi_effective_2019, jayaram_scaling_2020} have significantly improved their performance to begin to approach the locomotion capabilities of their larger counterparts \cite{st_pierre_toward_2019, de_croon_insect-inspired_2022}.
Despite this promising development, they are still limited by their largest body dimension for crevice traversal \cite{rao_analysis_2022}. 
To be effective in such situations, it is not sufficient to be just small, they need to be capable of shape manipulation \cite{murphy_marsupial_2000}.
Miniaturized soft robots are an obvious solution to tackle this problem due to their body compliance \cite{rus_design_2018}. 
However, their reliance on material properties only to achieve compliance limits their overall performance in terms of speed, payload, and onboard power \cite{hawkes_hard_2021}. 
An alternative approach is to take advantage of robot body geometries to achieve shape deformation inspired by observations of insect exoskeletons that are tough yet compliant \cite{jayaram_cockroaches_2016}.

\begin{figure} [!tb]
	\centering
	\includegraphics[width=\linewidth]{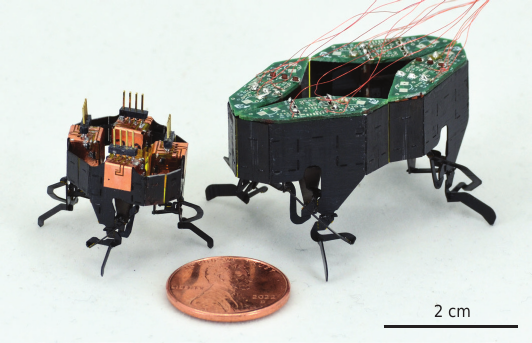}
	\caption{mCLARI (left) alongside the previous larger version CLARI (right). mCLARI measures a neutral body length of 20 mm, and weighs 0.97 g.}
	\label{fig:RobotOverview}
\end{figure}

\begin{figure} [!tb]
	\centering
	\includegraphics[width=\linewidth]{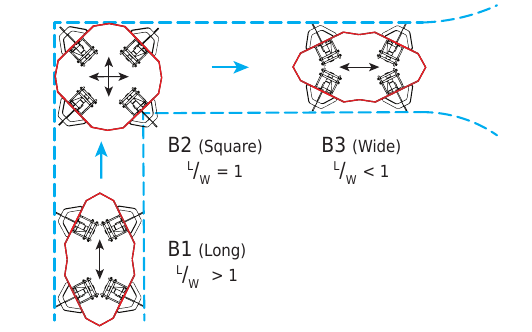}
	\caption{mCLARI body compliance configurations under environmental constraints. Robot body shape aspect ratios and directions of motion under compression. Neutral body length \SI{20}{\milli \meter}, with a maximum length of \SI{24}{\milli \meter} and width \SI{16}{\milli \meter} body shape combination.}
	\label{fig:RobotCompliance}
\end{figure}

Leveraging this embodied morphological intelligence principle \cite{sitti_physical_2021, nguyen_adopting_2022}, our group has previously demonstrated the first example of laterally confined locomotion \cite{mulvey2023deformobot} in legged robots using CLARI (Compliant Legged Ambulatory Robotic Insect) \cite{kabutz_design_2023}, an \SI{2.56}{\gram} insect-scale quadrupedal robot with a nominal body length of \SI{34}{mm}. 
This robot achieved a maximum speed of \SI{28}{\milli ms^{-1}} and is capable of locomotion in multiple shape configurations using a variety of gaits (trot, walk, etc.) and running frequencies (\SI{1}{}- \SI{10}{\hertz}). 
Forward locomotion performance was strongly correlated with the 'constrained' body shape - long and narrow (B1 from Fig. \ref{fig:RobotCompliance}) significantly faster than the short and wide configuration (B3 from Fig. \ref{fig:RobotCompliance}). 
Without effective tuning of compliance, the unconstrained robot simply oscillated in place, and no forward locomotion was observed.

In this paper, we improve on CLARI by presenting mCLARI (or mini CLARI, Fig. \ref{fig:RobotOverview}), a miniaturized version (\SI{60}{\%}) of its predecessor and yet significantly better performing in two primary aspects. 
First, the robot can locomote effectively without needing to 'fix' the body shape in open environments and demonstrates the capability of rapid laterally confined locomotion by passively adapting to its external constraints (Fig. \ref{fig:RobotCompliance}). 
mCLARI improved speed and performance in both unrestricted and confined environments, increasing speed to \SI{60}{\milli ms^{-1}}.
As a new capability, the robot is now omnidirectional within a confined space and can move forward or sideways with equal ease, drastically increasing its planar maneuverability (Fig. \ref{fig:RobotCompliance}). 
In the following sections, we will first highlight the morphological intelligence principle enabling confined terrrain locomotion and describe its embodiment in the mCLARI's exoskeleton similar to that of insects. 
We then detail the changes to the robot design compared to CLARI and note the implications of scaling followed by the characterization of the actuators, leg modules, and the whole robot itself.
We then demonstrate the new capability of mCLARI -- orthogonal confined terrain locomotion -- and experimentally quantify the performance. 
Finally, we conclude by contextualizing our work with respect to state-of-the-robots of similar sizes and highlight potential future improvements.

\section{Morphological Intelligence embodied in exoskeleton compliance}
Studies of crevice traversal in cockroaches revealed that tough yet deformable exoskeletons enable cockroaches to rapidly squeeze through vertically confined gaps as small as the height of two stacked pennies (\SI{3.2}{\milli m}) requiring body compression of over \SI{50}{\%} \cite{jayaram_cockroaches_2016}. 
By translating these insights into a bioinspired robot, CRAM, capable of vertically confined locomotion, the authors revealed passive compliance embodied in morphology as a mechanism of physical intelligence \cite{sitti_physical_2021, nguyen_adopting_2022}. 
Related studies indicate that mechanically tuned bodies are also effective in mitigating the effect of collisions and functioning as passive controllers that enable rapid transitions \cite{jayaram_transition_2018}. 
The use of exoskeleton-based compliance allows for increased degrees of freedom of the body shape compared to rigid body designs, while strategically limiting and controlling them when originating from soft material-based compliance. 
This allows for deliberate and passive local shape control constraining deformation to specified body regions.
Inspired by these biological observations, we embody morphological intelligence into the exoskeleton of mCLARI through design choices that utilize the geometry of the exoskeleton rather than soft material properties to achieve body compliance.
We realize this innovation in mCLARI through modular single leg segments that are connected in a closed kinematic chain to each other.

\begin{figure} [tb]
	\centering
	\includegraphics[width=\linewidth]{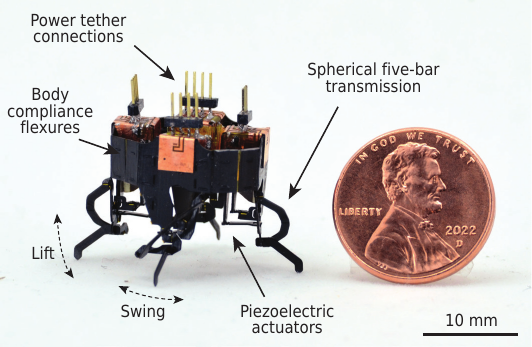}
	\caption{mCLARI system shown with a penny (19 mm diameter) for scale. Each of the leg modules is equipped with an individual pin header for wire connections. }
	\label{fig:mCLARI}
\end{figure}

\section{mCLARI: Design, Fabrication and Characterization}

\subsection{Platform Overview}
mCLARI (Fig. \ref{fig:mCLARI}) has a symmetric square body shape (B2, Fig. \ref{fig:RobotCompliance}) nominally with a body length of \SI{20}{mm}.
The robot is capable of compressing down to \SI{66}{\%} and expanding to \SI{150}{\%}. 
The total body mass is \SI{0.976}{g} including the carbon fiber body (\SI{0.188}{g}), eight piezoelectric actuators (\SI{0.544}{g}) and their frames (\SI{0.180}{g}), and miscellaneous assembly components such as glue, solder, etc. (\SI{0.064}{g}). 
The robot has four independently actuated leg modules, each with two degrees of freedom driven by piezoelectric actuators.
A spherical five-bar (SFB) linkage is used as the transmission that interlinks the motion of the two actuators in a single module to the output leg \cite{goldberg_gait_2017}.
Interconnecting the leg modules are laminates comprising of a single long flexure, which
enable planar motion for lateral shape deformation.
The robot was fabricated using the laminate manufacturing approach \cite{sreetharan_monolithic_2012} identical to CLARI \cite{kabutz_design_2023}. 
The current version of mCLARI is tethered for power and control.
For ease of testing, pin headers were soldered to the top of each leg module (Fig. \ref{fig:mCLARI}) which 
added $380 mg$ of payload during all robot experiments.

\begin{figure} [tb]
	\centering
	\includegraphics[width=0.9\linewidth]{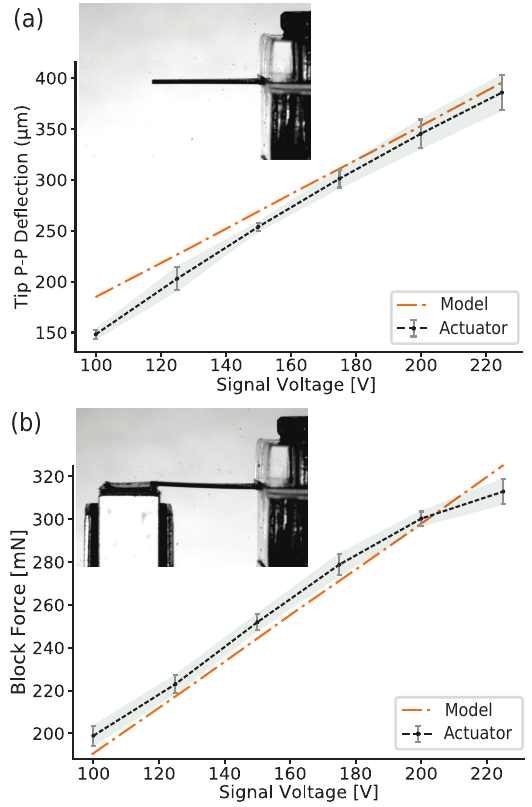}
	\caption{Characterization of mCLARI bimorph piezoelectric actuators (a) Free tip deflection of piezoelectric actuators compared to theoretical math model. (b) Single sided block force of piezoelectric actuators compared to theoretical math model. } 
	\label{fig:RobotActuatorChar}
\end{figure}

\subsection{Scaling Down Flexures, Actuators \& Transmissions}
The \SI{60}{\%} size reduction from CLARI \cite{kabutz_design_2023} to mCLARI required a significant redesign of the linkage portions due to performance considerations.
Fore instance, the miniaturization also forced the linkage inputs to the SFB to be located on the same plane as the main joints from the swing and lift which connect the transmission to the body ground frame.
Therefore, the input joints segments are now double layer folded to connect partially through the front layer, ensuring the planar joint alignment, minimizing the effects of directional biasing between lift and swing.
In general, the internal leg transmission pieces in mCLARI were scaled down to \SI{60}{\%} of their dimension in CLARI, up to a minimum flexure width of \SI{1}{\milli m} for reliability of the manufacturing.
However, the transmission flexure joints in mCLARI were increased in stiffness (\SI{\sim 4.6}{\times} by increasing the flexure layer thickness from \SI{7.5}{\micro m} to \SI{12.5}{\micro m}) to improve the mechanical lifetime. 
Additionally, the flexure length was also reduced to \SI{20}{\micro m} and now uses a mirrored castellation pattern \cite{doshi_model_2015}.

The actuators in mCLARI were reduced to \SI{\sim 80}{\%} relative to CLARI to maximize the power and locomotion performance of the robot. 
This effectively reduced their weight by \SI{\sim 50}{\%}, but retained $65\%$ of the displacement and produced a similar force as the CLARI actuators (Fig. \ref{fig:RobotActuatorChar}(a)).
With this improvement, we estimate mCLARI's payload body ratio ($r_{pl}$), defined as the maximum total load the robot can hold while maintaining its ability to walk, to be $r_{pl}=4.51$.  
To account for the increased relative actuator force, the mechanical ground of the leg modules was reinforced with a double layer fold at the location of lift and swing input to the SFB.
Furthermore, the actuator frame was extended to support the transmission point as a perpendicular rib, providing significant additional strength to the leg module.

\subsection{Actuation Characterization}
The actuator design and laminate layer stack up is the same as was used for CLARI \cite{kabutz_design_2023} \cite{jafferis_streamlined_2021}.
The mCLARI actuators' performance in terms of free deflection and blocked force is shown as a function of drive voltage levels in Fig. \ref{fig:RobotActuatorChar}.
At $225V$ sinusoidal peak-to-peak voltage signals, the actuators achieved on average $405 \mu m$ in free tip deflection (Fig. \ref{fig:RobotActuatorChar}(a)). 
The actuator's blocked force was measured at the tip with a FUTEK load cell, seen in the image in the top left corner (Fig. \ref{fig:RobotActuatorChar}(b)).
At $225V$, each actuator could support over \SI{30}{\times} the robot's body weight allowing significant mechanical advantage for the transmission. 
Our experimental results closely match the predictions of the mathematical model for both the free displacement and the block force \cite{jafferis_design_2015}.


\begin{figure} [htb]
	\centering
	\includegraphics[width=0.95\linewidth]{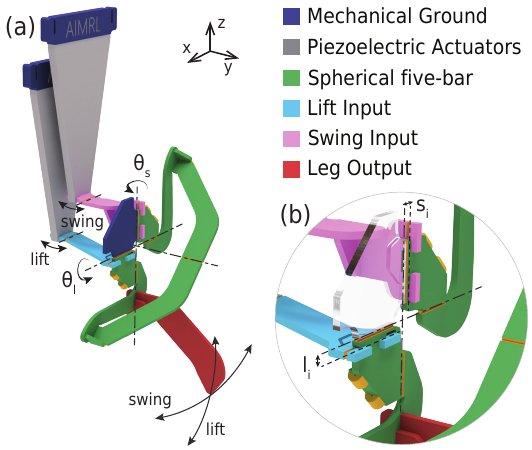}
	\caption{Mechanism of single leg module. (a) Actuator connecting to the leg tip across a spherical five-bar linkage. (b) Magnified view of the leverage distances for the input of actuation.}
	\label{fig:RobotLeg}
\end{figure}

\subsection{Transmission Characterization}

\begin{figure} [tb]
	\centering
	\includegraphics[width=0.9\linewidth]{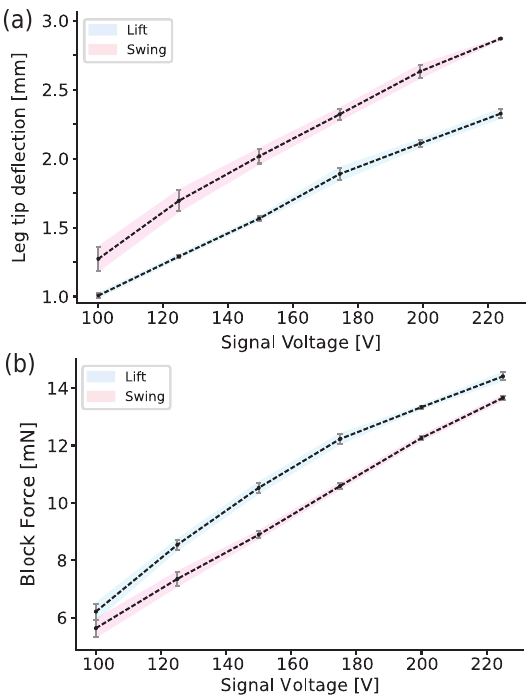}
	\caption{Characterization of mCLARI transmission mechanics (a) Leg tip free deflection in Lift and Swing directions. (b) Leg tip block force in Lift and Swing directions.} 
	\label{fig:RobotLegChar}
\end{figure}

To improve the SFB transmission (Fig. \ref{fig:RobotLeg}) output performance by limiting plastic deformation due to off-axis forces, both the swing and lift lever arms ratios were reduced relative to CLARI. 
The lift transmission ratio ($T_{Ratio,L} = \frac{l_d}{l_i}$), where $l_d$ and $s_d$ is the distance from the tip of the leg to the center of the transmission joint for lift and swing respectively, was designed at $T_{Ratio,L} = 12.5$, with the lift input connection to the SFB being $l_i = 500 \mu m$. 
The lift is biased towards having larger force and leg stiffness to support the body mass and additional payload.
The leg tip is calculated to be placed $6.25mm$ below the location of the transmission center joint.
The transmission ratio in the swing direction ($T_{Ratio,S} = \frac{s_d}{s_i}$) was designed to $T_{Ratio,S} = 10$.
The leg length was reduced to $4.5mm$ from the transmission center joint location to allow robot navigation through lateral constraint spaces with minimal interference to the side. 
Thus, an swing input connection distance of $s_i = 450 \mu m$ was used in the transmission to the SFB.

The free tip position range of the leg tip was tested at various signal voltages  (Fig. \ref{fig:RobotLegChar}(a)) in both the lift and the swing directions.
As desired, the swing deflection was consistently more than the lift deflection, with a maximum of $2.85 mm$ leg tip swing at $225V$, when powered at low frequency sinusoidal peak-to-peak voltage signals in the quasi-static region.
The lift achieved $2.3 mm$ at $225V$ without load, providing enough range to have leg lift under body load and external payload.
Both the lift and swing displacements from mCLARI are equivalent to what the previous CLARI robot achieved with the larger body scale.
The leg tip blocked force (Fig. \ref{fig:RobotLegChar} (b)) was measured to be higher in the lift direction than the swing direction as designed.
At $225V$, the lift block force is $14.3 mN$, making it possible for a single leg to carry the full robot body weight.

\subsection{Modular leg mechanics}
The modularity of the mCLARI leg modules was improved by redesigning the electrical circuit frame of the actuator side wall.
An origami-style foldable flexible electronics board was developed to provide the structural integrity required to hold the actuators in place to the leg transmission, and have flexible joints with conductive traces to ease the assembly process of a leg module. 
The structure of the actuator frame layer is FR4 fiberglass ($127 \mu m$) as the structural layer, FR1500 ($13\mu m$) as adhesive and a copper-clad Kapton prestack of polyimide ($25 \mu m$), adhesive ($25 \mu m$) and copper ($17.5 \mu m$).
The rigid layer is only cut into shapes for the side walls and structural components with tabs on the side to interlock with the carbon fiber leg transmission. 
Using our custom laser micromachine (6D Lasers), copper is rastered away to leave traces on the bare kapton film to connect central tether wires to each of the piezoelectric actuator pins. 
For the actuator frame, the folding joints were $300 \mu m$ wide, allowing for a folding radius that results in minimal strain tearing of the copper layer.
The frame traces connect 4 pins from a pin header (each connected to a shared ground, a shared high voltage bias, a lift voltage signal and a swing voltage signal, respectively) to the top of the leg module to allow for ease of testing.

\begin{figure} [!htb]
	\centering
	\includegraphics[width=0.9\linewidth]{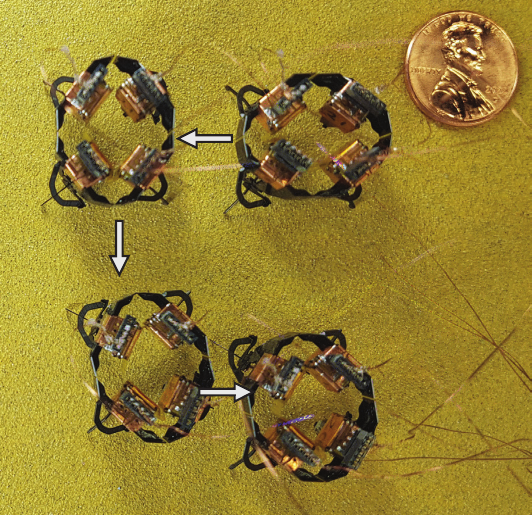}
	\caption{mCLARI demonstrating omnidirectional movement. Starting in the top right corner, walking to the left, then without turning walking downwards,followed by movement to the right.}
	\label{fig:RobotOmni}
\end{figure}

\section{Robot Performance}
In this section, we demonstrate mCLARI's novel abilities of omnidirectionality and confined terrain locomotion and measure its performance.

\subsection{Experimental setup}
A custom script running \texttt{MATLAB Simulink} generateed the control signals required to drive the robot through a Speedgoat real-time operating computer.
These signals are boosted up $100 \times$ through a custom built voltage amplifier and transmitted to the robot via the 6-strand magnet wire assembly (48 gauge).

\subsection{Unconstrained Locomotion}
To characterize the unconfined locomotion performance, the robot was run on different surfaces, namely cardboard, paper, sandpaper (220, 180, 120, 80, 60 grit roughness). 
The best robot performance was observed with 120-grit sandpaper, which was subsequently used as a ground surface for confined locomotion.
mCLARI achieves a top speed of \SI{60}{\milli ms^{-1}} (\SI{\sim 3}{bodylengths/s}) at a frequency of \SI{10}{\hertz} with the classical trot gait \cite{goldberg_gait_2017}.

\begin{figure} [!htb]
	\centering
	\includegraphics[width=\linewidth]{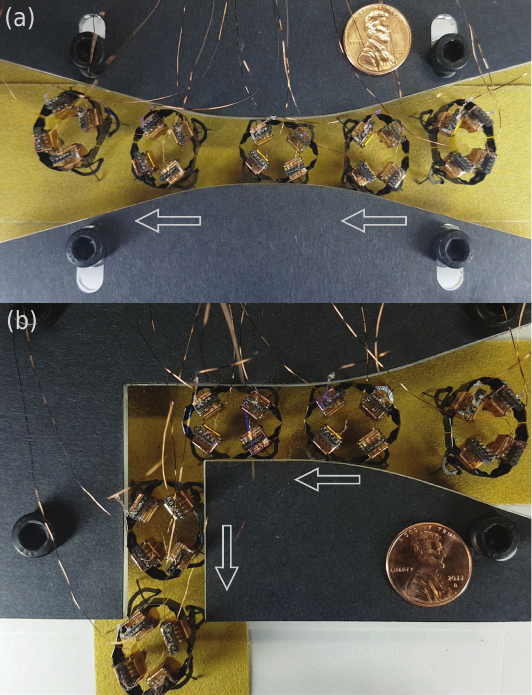}
	\caption{mCLARI passive shape morphing demonstration. (a) mCLARI walking from right to left through a narrow lateral constraint. (b) mCLARI walking through a lateral constraint with a 90\degree bend, which is too tight for turning. The robot moves from right to left in long configuration, then from top to bottom in the wide configuration.}
	\label{fig:ComplianceExperiments}
\end{figure}

\subsection{Omni-directionality}
For all omnidirectional runs, the same classic trot gait was used, with only inter leg phase shifts being applied to change the robot walking direction.
The locomotion performance was observed to be similar forward and laterally, with the effect of a single 'weaker' leg being observed experimentally as directional walking drift in videos.
The novel ability of omnidirectionality, specifically orthogonal motion without body rotation, is demonstrated in Fig. \ref{fig:RobotOmni}, where mCLARI starts in the top right corner, then walks towards the left, followed by walking downwards without turning, and then toward the right.

\subsection{Confined terrain navigation}
The body compliance of mCLARI enables the robot to maneuver through confined terrain smaller than its lateral width in its nominal shape.
In Fig. \ref{fig:ComplianceExperiments}(a), the robot trots in its symmetric starting shape ($20 mm$) towards the confined spacing gap of $16.5mm$. 
As the robot approaches the gap, the external forces on the body deform the shape to adapt to its environment.
Once the robot has squeezed through the confinement, the passive compliance returns the body it's neutral nominal shape.
Note that all the confined space experiments restricted only the body, leaving space for the legs to move freely without interference. 

\subsection{Complex Cluttered Terrain Locomotion}
Combining the ability of omnidirectionality and compliance in confined terrain, mCLARI has demonstrated the ability to walk through a $90 \degree$ confinement corner, the first in any legged robot.
Attempting to turn like conventional robots would get it stuck in the tight corner of the confined terrain. 
Instead, mCLARI leverages its ability of omnidirection compliance to change the direction of motion and continue sideways in the next compressed body shape.
In Fig. \ref{fig:ComplianceExperiments}(b) the robot starts in a free body shape on the right, then trots to the left into a body confinement of $16.5 mm$. 
In the constrained environment, the robot does not have sufficient space to turn, with the only option to move downward into a new compressed body shape through omnidirectional motion.

\begin{table}[h]
    \centering
    \caption{mCLARI compared against other microrobots}
    \label{tab:Robot_comparison}
    \begin{tabular}{p{1.7cm}|p{0.6cm}|p{0.6cm}|p{0.6cm}|p{0.6cm}|p{0.6cm}|p{0.6cm}}
        \hline
        ROBOT & Body length [$mm$] & Rel. Speed [BLs$^{-1}$]& Omni directional & Soft robot & Lateral compliance & Payload ($\times$ body weight)\\
        \hline
        \hline
        mCLARI & 20 & 3 & Yes & Yes & Yes & 4.51\\
        CLARI \cite{kabutz_design_2023} & 34 & 0.8 & No & Yes & Yes & 2\\
        HAMR-Jr \cite{jayaram_scaling_2020} & 22.5 & 13.9 & No & No & No & 10\\
        Wu \cite{wu_insect-scale_2019} & 10 & 20 & No & Yes & No & 6\\
        St.Pierre \cite{pierre_3d-printed_2018} & 2.5 & 14.9 & No & No & No & -\\
        Shin \cite{shin_micro_2012} & 20 & 2.4 & No & No & No & -\\
        Kim \cite{kim_5_2019} & 2 & 4 & No & No & No & -\\
        \hline
    \end{tabular}
\end{table}

\section{Conclusion and Future Work}

To conclude, we successfully designed and fabricated an insect-scale miniature compliant robot capable passive shape morphing to locomote in confined terrains. 
We improved on its predecessor by reducing body size to $60\%$, mass to $38\%$ of the original, while maintaining $80\%$ of the actuation power.
mCLARI demonstrates for the first time -- omnidirectional, lateral confined space locomotion capability, a unique capability among legged robots (Tab. \ref{tab:Robot_comparison}).
Unlike, previous studies \cite{schiebel_passive_2022, lathrop_directionally_2023} which leveraged morphological intelligent mechanisms embodied in the legs to extend locomotion performance in rough and laterally confined terrains, mCLARI showcases the ability for passive shape adaptation to environmental constraints and represents the first step towards highly functional shape morphing systems that can potentially enable unprecedented access to new complex environments.
As immediate next steps for mCLARI, we aim to integrate controllable body compliance and ground contact mechanisms for enabling active shape-shifting \cite{shah2021shape} in miniature robots. 
In the long term, we aim to add onboard closed-loop control \cite{jayaram_concomitant_2018}  and autonomous decision making to enable systems like mCLARI to go where no robots have gone before.





\section*{ACKNOWLEDGMENT}

We thank all the members of the Animal Inspired Movement and Robotics Lab at the University of Colorado Boulder for their valuable support and suggestions with the robot design and testing. We also thank Stephen Uhlhorn and 6DLaser for the femtosecond laser micromachine development and fabrication assistance.


\medskip

\bibliographystyle{ieeetr}
\bibliography{IROS_mCLARI.bib}

\end{document}